\documentclass[10pt,twocolumn,letterpaper]{article}

\usepackage[pagenumbers]{cvpr} 

\usepackage[dvipsnames]{xcolor}

\definecolor{cvprblue}{rgb}{0.21,0.49,0.74}
\usepackage[pagebackref,breaklinks,colorlinks,citecolor=cvprblue]{hyperref}
\usepackage{graphicx}
\usepackage{amsmath}
\usepackage{amssymb}
\usepackage{booktabs}
\usepackage{multirow}
\usepackage{amsfonts}
\usepackage{algorithm}
\usepackage{algorithmic}
\usepackage{bm}
\usepackage[dvipsnames]{xcolor}
\usepackage{colortbl} 
\usepackage{xcolor}

\newcommand{\authorskip}{\hspace{5mm}}

\title{\textit{VIDiff}: Translating \underline{\textit{V}}ideos via Multi-Modal \underline{\textit{I}}nstructions with \underline{\textit{Diff}}usion Models}

\author{Zhen Xing\textsuperscript{1} \authorskip~~Shuyuan Tu\textsuperscript{1}\authorskip Qi Dai\textsuperscript{2} ~~Zihao Zhang\textsuperscript{1}\authorskip Hui Zhang\textsuperscript{1}  \authorskip Han Hu\textsuperscript{2}  \authorskip Zuxuan Wu\textsuperscript{1} \authorskip Yu-Gang Jiang\textsuperscript{1} \\[0.5mm]
{\textsuperscript{1} Fudan University} 
~~{\textsuperscript{2} Microsoft Research Asia}
}

\begin{document}
\twocolumn[{
\maketitle
\vspace{-1.8em}
\renewcommand\twocolumn[1][]{#1}
\begin{center}
    \centering
    \includegraphics[width=1.0\textwidth]{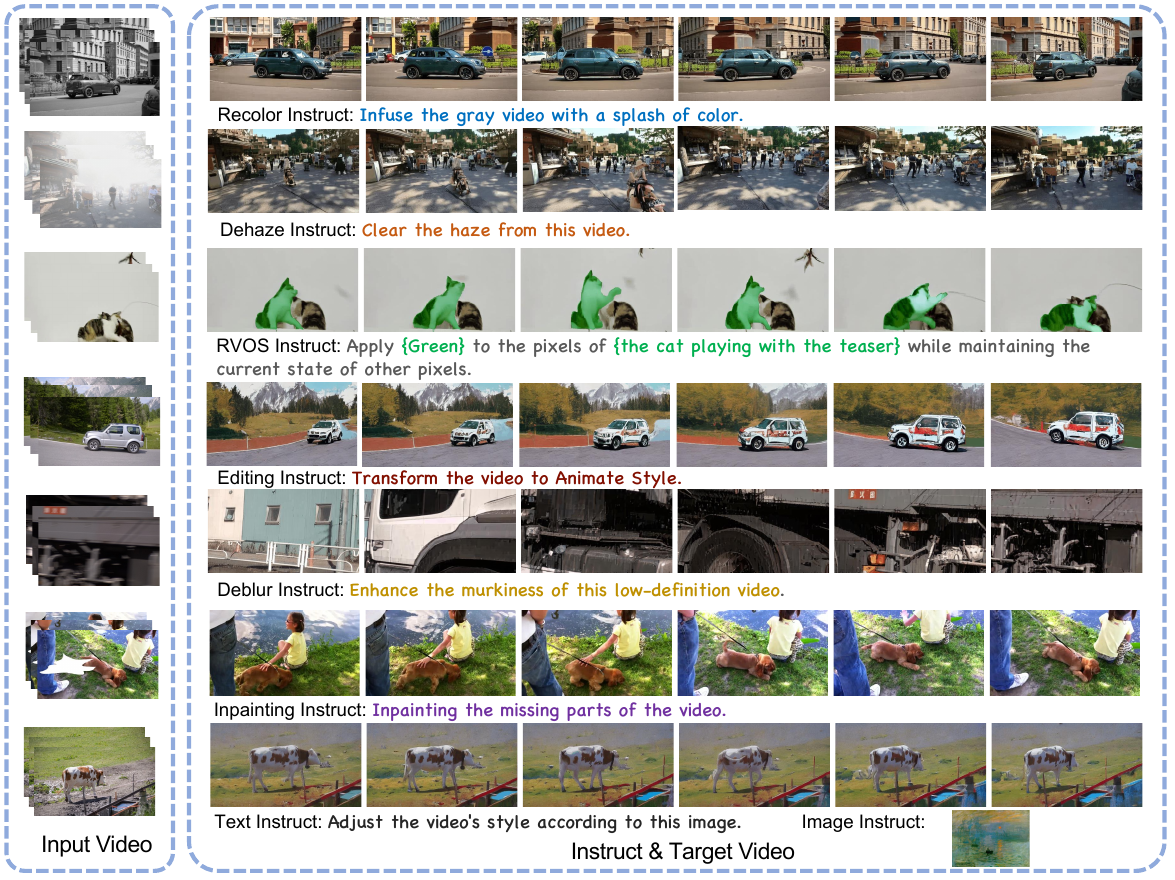}
    \vspace{-0.5cm}
    \captionof{figure}{
    We introduce VIDiff, a generalist model for video translation tasks. Given an input video and human instructions, our unified model effectively accomplishes tasks such as video re-colorization, dahazing, deblurring, editing, in-painting, and object segmentation.
    }
    \label{fig:fig1}
    \vspace{-0.0cm}
\end{center}
}]

\begin{abstract}
\vspace{-0.3cm}

Diffusion models have achieved significant success in image and video generation. This motivates a growing interest in video editing tasks, where videos are edited according to provided text descriptions. However, most existing approaches only focus on video editing for short clips and rely on time-consuming tuning or inference. 
We are the first to propose Video Instruction Diffusion (VIDiff), a unified foundation model designed for a wide range of video tasks.  
These tasks encompass both understanding tasks (such as language-guided video object segmentation) and generative tasks (video editing and enhancement). 
Our model can edit and translate the desired results within seconds based on user instructions. 
Moreover, we design an iterative auto-regressive method to ensure consistency in editing and enhancing long videos. We provide convincing generative results for diverse input videos and written instructions, both qualitatively and quantitatively. More examples can be found at our website \url{https://ChenHsing.github.io/VIDiff}.
\end{abstract}

\vspace{-0.7cm}
\section{Introduction}
In recent years, the field of artificial intelligence has witnessed significant advancements, especially in Natural Language Processing (NLP), where Large Language Models (LLMs) like GPT~\cite{GPT4} unify diverse tasks under one single framework. 
In contrast, the development of foundational models in computer vision is still far behind that in NLP, due to the natural diversities arising from vision tasks, \emph{e.g.}, various output formats, and different model architectures.

Inspired by the success of GPT~\cite{GPT4} in unifying NLP tasks, some foundational models have emerged that aim to unify visual tasks, 
primarily focusing on understanding tasks like recognition and retrieval~\cite{omnivl, blip2, beit3}. 
Nonetheless, research on unified frameworks for generative tasks is relatively scarce. 
InstructDiffusion~\cite{instructdiffusion} explores generalizing diffusion models to both image editing and understanding tasks. 
Despite that, unifying tasks in the video domain is still challenging, since the data distribution and task variation are more complex than images.
Few have endeavored to design a unified framework that addresses both video understanding and editing tasks. 

Among generative modeling tasks~\cite{hertz2022prompt2prompt, brooks2023instructpix2pix,stablediffusion, SimDA,vdmsurvey, adadiff, ssl3d, mpcn},
Video-to-Video (V2V) translation possesses enormous potential in social media, advertising, promotional campaigns, television, \emph{etc}. 
Presently, most methods rely on detailed textual descriptions, a strict requirement hinging on accurate descriptions of the original and target videos. In addition, the majority of methods depend on time-consuming training and inference processes such as DDIM~\cite{ddim} inversion. 
While instructional editing~\cite{brooks2023instructpix2pix,qin2023instructvid2vid,InsV2V, gan2023instructcv} takes in user-friendly prompts,  current techniques can only be applied to a very few editing scenarios.
Besides, while instructional texts are able to relieve the need for professional prompts, producing precise and detailed descriptions of expected outputs sometimes require domain knowledge, \eg, art or medicine. It is desirable to provide more effective instructions to enable effortless guidance.

To address these concerns, in this paper, we present a general video diffusion framework, VIDiff, for various conditional video-to-video translation that operates on multimodal instructions.
Our method accepts instructions together with a source video as input and generates a target video output. 
In addition to the textual instruction, we also leverage images that ``worth a thousand words'' as straightforward instruction without demanding expert knowledge.
We therefore design a multi-modal condition injection mechanism for image and text-guided video editing.
Our approach is trained with multiple stages to adapt a pre-trained T2I model~\cite{stablediffusion} for V2V translation.
We also design an iterative training and inference scheme to allow long video translation. 
We effortlessly extend our method to various tasks, building up a unified framework for video understanding and editing.

In summary, the main contributions of this paper can be summarized as follows:
\begin{itemize}
    \item We are the first to employ a unified diffusion framework for both video understanding and video enhancement tasks.
    \item We design a multi-stage training method to seamlessly transfer the T2I model for multi-modal conditional video translation tasks.
    \item Our proposed iterative generation method is simple yet effective, allowing easy application in long video translation tasks.
    \item We conduct extensive experiments, showcasing several cases, proving the effectiveness of our approach both qualitatively and quantitatively.
\end{itemize}

\section{Related Work}
\noindent\textbf{Video Language Foundation Model.}
Although image-language foundation models have been successfully applied to various tasks, including image recognition~\cite{clip}, image-text retrieval~\cite{clip}, visual question answering~\cite{vqa}, and even image generation and editing~\cite{instructdiffusion}, there has been limited research on video-language foundation models~\cite{videollm, openvclip, openvclip2}. Existing methods are typically designed for understanding tasks like classification. Inspired by contrastive learning, Omnivl~\cite{omnivl} explores cross-modal alignment for images, text, and videos, demonstrating effectiveness in video classification and retrieval tasks. Unmasked Teacher~\cite{unmaskedteacher} combines masked auto-encoder with contrastive learning in a multimodal paradigm, making it applicable to diverse video-language tasks such as classification, retrieval, temporal detection, and video question-answering.
Approaches like Unicorn~\cite{unicorn} and OmniTracker~\cite{omnitracker} aim to unify video object segmentation and tracking tasks. However, there has been limited research on video translation tasks. We are the first to design multiple video translation tasks into a unified foundation model.

\vspace{0.05in}
\noindent\textbf{Text-guided Image Translation.}
Image editing is a complex process that involves modifying an image based on specific guidance, often provided by a reference image, rather than generating images without constraints. Various methods have been developed to address this task. One approach includes zero-shot image-to-image translation techniques, like SDEdit~\cite{meng2021sdedit}, which applies diffusion and denoising techniques to a reference image. Other methods incorporate optimization techniques to refine the editing process. For example, Imagic~\cite{imagic} utilizes textual inversion concepts from ~\cite{imageinversion}. Null-text Inversion~\cite{Nulltext} leverages  Prompt-to-Prompt~\cite{hertz2022prompt2prompt} to control cross-attention~\cite{li2023transformer} behavior in the diffusion model. However, these methods require a time-consuming editing process due to the need for per-image optimization. Instead,
Instruct Pix2Pix~\cite{brooks2023instructpix2pix} achieves image editing by training on paired synthetic data.  More recently, InstructDiffusion~\cite{instructdiffusion} unifies several vision tasks under this paradigm. In this paper, we focus on the video translation task, which is more challenging compared to images.

\vspace{0.05in}
\noindent\textbf{Text guided Video Editing.}
Video editing methods often require detailed textual descriptions of both the original and target videos, and then reconstruct the videos based on these descriptions for editing purposes. Tune-A-Video~\cite{tuneavideo} and SimDA~\cite{SimDA} fine-tune a single model to generate new videos with similar motion patterns. Video-P2P~\cite{liu2023videop2p}, Vid2Vid-Zero~\cite{vid2vid-zero}, and FateZero~\cite{qi2023fatezero} leverage cross-attention maps to adjust videos. More recently, InstructVid2Vid~\cite{qin2023instructvid2vid} and InsV2V~\cite{InsV2V} attempt to build instruction-based video editing. While we share a similar structure, our focus is on unifying various video tasks in a generalist framework, and our approach is also applicable to editing long videos.

\section{Method}
In this section, we first introduce the preliminaries of the latent diffusion model (LDM) in Sec.~\ref{Sec:preliminary}. Next, we explain the problem definition in Sec.~\ref{Sec:probledef}. Then, we present the collection of the dataset used in our approach in Sec.~\ref{Sec:dataset}. Finally, we describe the architecture and training pipeline of our VIDiff in Sec.~\ref{Sec:unified}. 

\subsection{Preliminaries of LDM}
\label{Sec:preliminary}
Diffusion models~\cite{ddpm, ddim}  model complex data distributions by two pivotal processes: diffusion and denoising. Given an input data sample $\bm{x}$ from the distribution $\bm{p}(\bm{x})$, the diffusion process adds random noise to transform the sample to $\bm{x}_t={\alpha}_t \bm{x} + {\sigma}_t\bm{\epsilon}$. where $\bm{\epsilon}$ is sampled from a standard normal distribution $\mathcal{N}(\bm{0},\mathbf{I})$. This diffusion process is achieved by $T$ steps, and the noise scheduler is parametrized by the parameters ${\alpha}_t$ and ${\sigma}_t$. In the denoising stage, the model employs $\bm{\epsilon}$-prediction and $\bm{v}$-prediction methodologies to learn a denoiser function $\bm{\epsilon}_{\theta}$, which is trained to minimize the mean square error loss as follows:
\begin{equation}
\mathbb{E}_{\bm{x},\bm{\epsilon} \sim \mathcal{N}(\bm{0},\mathbf{I}),t}[\lvert\lvert \bm{\epsilon} - \bm{\epsilon}_\theta(\bm{x}_t,t)\rvert\rvert^2_2].
\end{equation}
Latent Diffusion Model (LDM)~\cite{stablediffusion} utilizes a VAE~\cite{VQVAE} encoder $\mathcal{E}$ to compress the input data in low-dimensional latent space. LDM conducts diffusion and denoising processes in both the training and inference stages. The optimizing objective is:
\begin{equation}
\mathbb{E}_{\bm{x},\bm{\epsilon} \sim \mathcal{N}(\bm{0},\mathbf{I}),t}[\lvert\lvert \bm{\epsilon} - \bm{\epsilon}_\theta(\mathcal{E}(\bm{x}_t),\bm{c},t)\rvert\rvert^2_2],
\end{equation}
where $\bm{c}$ is the text condition that is extracted by the pretrained CLIP~\cite{clip} ViT-L/14 model from the text prompt. LDM is a text-to-image model, and we adapt it to video-to-video translation tasks in this work.

\subsection{Problem Definition}
\label{Sec:probledef}
Video understanding~\cite{videosum, wang2023look} and generative tasks differ in various aspects.
However, we can reformulate each task and raise some commonalities.
For most common video tasks, we can consider them as conditional video translation tasks. For instance, video object segmentation can be seen as translating raw video pixels into corresponding segmentation maps. Video recoloring task involves translating grayscale video pixels into colored video frames. As for video enhancement and video editing tasks, they are inherently video translation tasks as well.

We intend to design a unified model capable of addressing all these tasks simultaneously. Thus, we tackle the aforementioned tasks uniformly in an instructional video translation. 
Specifically, given a source video $V_s$ and an instruction $\bm{c}$, the objective is to translate $V_s$ into the corresponding target video $V_t$. To achieve this goal, during the training phase, we construct training video triplets $⟨V_s, V_t, \bm{c}⟩$ for each task. In the inference phase, the method could translate a source video $V_s$ to the target video $V_t$ conditioned on instruction $\bm{c}$.

\subsection{Training Data Construction}
\label{Sec:dataset}
As previously mentioned, the training of a video-to-video translation model relies on the construction of triplets, which consist of $<V_s, V_t, \bm{c}>$. In this section, we will discuss how to collect datasets for various tasks. The visualization of the triplet dataset can be found in Fig.~\ref{fig:fig1}.

\vspace{0.2cm}
\noindent\textbf{Video Re-colorization and Inpainting}
~~For tasks like video re-colorization and video inpainting, we can easily construct training data using unlabeled videos. Any video can be converted into a grayscale version, and videos with missing parts can be generated by creating masks of arbitrary shapes~\cite{propainter}. As for the instruction, we can write phrases like ``convert the grayscale clip into a colorful masterpiece" and ``repair the video with missing parts." This approach allows us to effortlessly obtain video triplets.

\vspace{0.2cm}
\noindent\textbf{Video Dehazing and Deblurring}
~~For enhancement tasks like video denoising and dehazing, we can utilize commonly used datasets~\cite{BSD, hazeworld} in this domain, all of which are annotated with corresponding input and ground-truth data. Therefore, we only need to write a few instruction phrases manually, such as ``remove the applied haze from this video" and ``enhance the clarity of this blurry video."

\vspace{0.2cm}
\noindent\textbf{Language-guided Video Object Segmentation}
~~For the language-guided video object segmentation task, the goal is to identify and segment objects within the video based on natural language instructions. We utilize established datasets specified for this task~\cite{refer_davis,refer_youtube,mevis} for training. As for the instruction, we can manually craft phrases such as ``change the \{object\} pixels to \{color\}, while keeping the other pixels constant." These kinds of instructions can yield superior visual results, as validated in~\cite{instructdiffusion}.

\vspace{0.2cm}
\noindent\textbf{Instruction-guided Video Editing}
~~For most diffusion-based video editing methods, detailed textual descriptions of both the source and target videos are required. Additionally, operations such as one-shot tuning~\cite{tuneavideo} during training and DDIM Inversion~\cite{ddim} at the inference stage are time-consuming and resource-intensive. For a short video, this process requires several minutes, limiting its practicality. In contrast, our approach only requires the original video and editing instructions during the inference stage, enabling video editing within seconds. Nevertheless, constructing video editing datasets is challenging. We  follow~\cite{brooks2023instructpix2pix, qin2023instructvid2vid} to utilize GPT~\cite{GPT4} and the excellent video editing models~\cite{yang2023rerenderavideo, tuneavideo, SimDA, ccedit} to create triplet training data.

\begin{figure}
    \centering
    \includegraphics[width=1.0\linewidth]{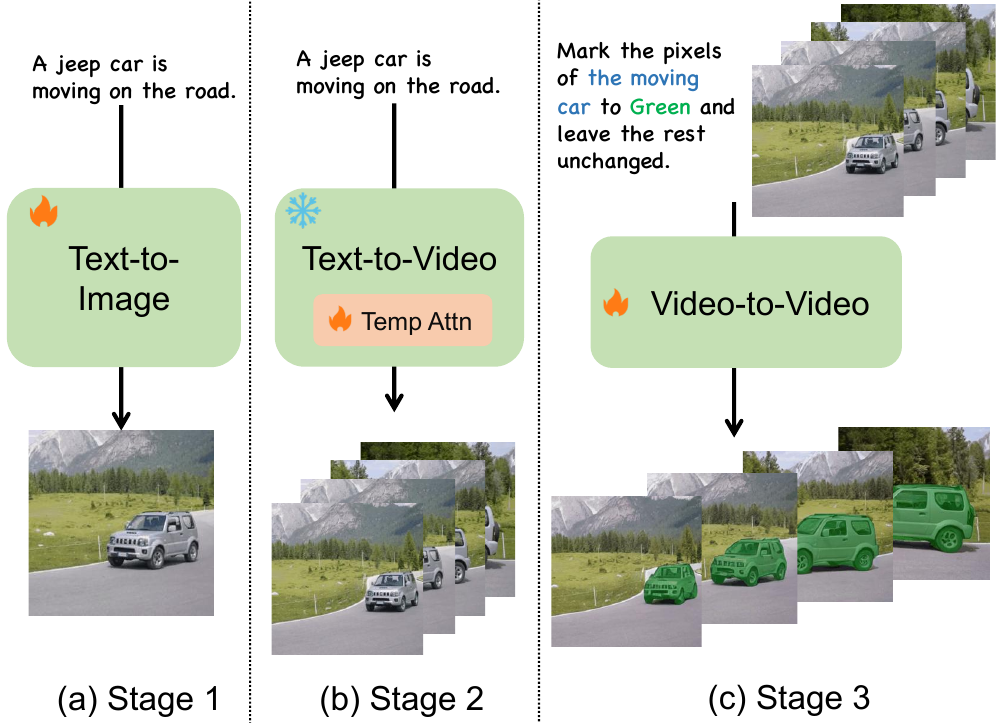}
    \vspace{-0.3cm}
    \caption{The overview of the training stage, which shows how to transfer a T2I model for V2V translation tasks. (a) Text-to-Image stage, (b) Text-to-Video stage, (c) Video-to-Video stage. }
    \label{fig:stage}
    \vspace{-0.3cm}
\end{figure}

\subsection{Unified Instructional Model for Video Tasks}
\label{Sec:unified}
In this subsection, we present how to design a unified instructional model to handle various video tasks and discuss how to transfer a pre-trained T2I model for general video translation tasks.


\vspace{0.2cm}
\noindent\textbf{Architecture}
~A common T2I model~\cite{stablediffusion} contains
a modified U-Net~\cite{unet} comprising 4 downsample/upsample blocks, and 1 middle block. Each block typically consists of spatial 2D convolutional layers, self-attention layers, and cross-attention layers with the text condition. To cope with video inputs, we inflate 2D convolutional layers into 3D convolutions~\cite{vdm}. Additionally, we add a vanilla temporal attention~\cite{vivit, timesformer, xing2023svformer} layer for motion modeling. Before passing a video with \texttt{f} frames $[\texttt{b, c, f, h, w}]$ to the temporal module, we reshape it into $[\texttt{(b h w), f, c}]$. To seamlessly integrate the temporal module into the training process without causing any adverse effects, we adopt a zero initialization approach for the output projection layer of the temporal transformer following~\cite{controlnet, AnimateDiff, SimDA}. 

\vspace{0.2cm}
\noindent\textbf{Training Stage}
~~As shown in Fig.~\ref{fig:stage}, we design a multi-stage training method to transfer a T2I model for V2V translation. 
The first stage is exactly the original T2I~\cite{stablediffusion} model training.
In the second stage, we introduce the aforementioned temporal attention layer and inflate the U-Net from 2D to 3D. By fixing the parameters of the original T2I model, we tune the temporal module to achieve T2V generation with a video-text dataset~\cite{webvid}. Leveraging pre-training from the previous stage, the model learns temporal motion modeling well. In the final stage, we fine-tune the pre-trained network using the collected datasets to accomplish the video-to-video translation task.

\begin{figure}
    \centering
    \includegraphics[width=1.0\linewidth]{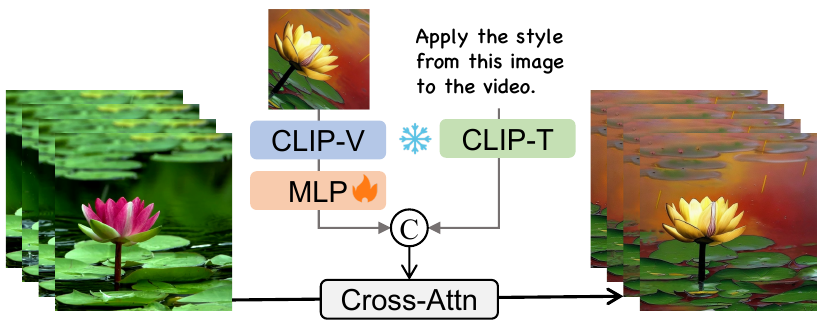}
    \vspace{-0.3cm}
    \caption{The overview of our multi-modal conditional method. During the training stage, we freeze the CLIP-Text and CLIP-Vision Encoder and only finetune the parameters of the MLP. }
    \label{fig:condition}
    \vspace{-0.3cm}
\end{figure}

\begin{figure*}
    \centering
    \includegraphics[width=0.95\textwidth]{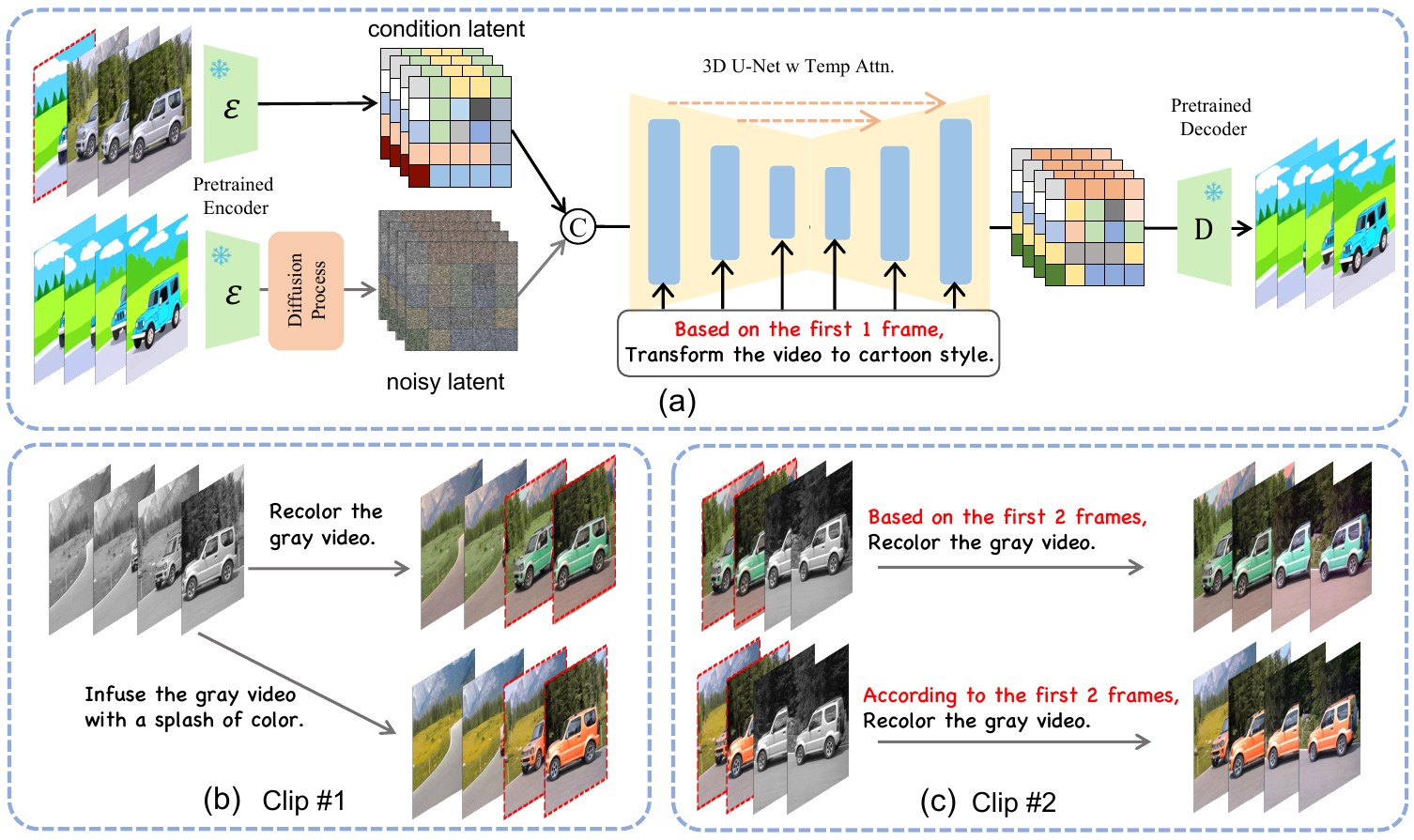}
    \vspace{-0.3cm}
    \caption{Pipeline of our VIDiff translation framework. We utilize the pre-trained auto-encoder as in Stable Diffusion~\cite{stablediffusion} to obtain latent representation. (a) During training, we randomly decide to use the first several frames as the condition. (b) The inference of the first clip. (c) Utilize the last frames of the video as the condition, we can iteratively translate the long videos.  }
    \vspace{-0.6cm}
    \label{fig:pipeline}
\end{figure*}

\vspace{0.2cm}
\noindent\textbf{Multi-Modal Condition Injection Mechanism}
~~Most previous video editing methods rely on provided textual descriptions or specific instructions~\cite{tuneavideo, liu2023videop2p, controlvideo,SimDA, ccedit}. Here, we introduce a straightforward multi-modal condition injection mechanism for video editing as shown in Fig.~\ref{fig:condition}. 
For a given textual instruction, we use the CLIP-Text~\cite{clip} Encoder to extract the embedding of the text. Additionally, we aim to incorporate images as visual instructions to learn editing patterns related to image styles. During training, we randomly select a frame from the target video and apply data augmentations such as flipping, rotation and cropping to create the image instruction. This image instruction is then processed through a pre-trained CLIP-Vision~\cite{clip} Encoder and a newly added MLP layer. Subsequently, we concatenate the resulting image embedding and the text embedding along the channel dimension to form a joint instruction embedding. In this training setting, the CLIP vision and text encoders remain fixed, and only the MLP layer needs training. This approach allows effective image instruction, which eliminates the professional text instruction that demands expert knowledge.

\vspace{0.2cm}
\noindent\textbf{Training pipeline}
~~The detailed training pipline for stage 3 is shown in Fig.~\ref{fig:pipeline} (a).
First, the source video $V_s$ and the target video $V_t$ are both input into a pre-trained VAE~\cite{VQVAE} encoder to transform them to $\bm{x}_s$ and $\bm{x}_t$ in latent space. Subsequently, noise is added to the target video through the diffusion process. The noisy latent, along with the latent of the source video (\emph{i.e.} condition latent), is concatenated in the channel dimension and fed into the inflated U-Net with temporal attention. During this process, the source video $\bm{x}_s$ and the instruction $\bm{c}$ serve as conditions to control the denoising process to translate to the target video $\bm{x}_t$. We minimize the following latent diffusion objective:
\begin{equation}
\mathbb{E}_{\bm{x}_s,\bm{x}_t,\bm{\epsilon} \sim \mathcal{N}(\bm{0},\mathbf{I}),t}[\lvert\lvert \bm{\epsilon} - \bm{\epsilon}_\theta([\bm{x}_t,\bm{x}_{s}],\bm{c},t)\rvert\rvert^2_2].
\end{equation}

In addition to translating short video clips based on instructions, our approach can also be extended to long video translation. The training process is also straightforward. When constructing training pairs, we randomly select the first $\bf{n}$ frames of the source video to align with the target video. Additionally, we modify the instruction to include phrases like ``Based on the first {$\bf{n}$} frames." This allows the network to learn the correspondence between temporal modeling and the frame number $\bf{n}$ specified in the instruction. Additionally, this approach facilitates the maintenance of coherence in the translation across each clip by leveraging information from the $\bf{n}$ reference frames when computing temporal attention.

During training, we combine multiple tasks into a unified training paradigm, randomly selecting a specific task at each training step. We will demonstrate the effectiveness of this multi-task training paradigm and the integration of diverse tasks within a single model in Sec.~\ref{Sec:ablationstudy}.

\vspace{0.1cm}
\noindent\textbf{Inference pipeline}
~~During inference, we employ an iterative inference approach for the long videos. As illustrated in Fig~\ref{fig:pipeline}(b), for the first clip $\#1$, we employ a regular inference approach. Based on the source video and instructions as conditions, the model progressively denoises from Gaussian noise to derive the target video. Once the first clip is obtained, we can use the last ${n}$ frames from clip $\#1$ as the condition. For the next clip $\#2$, we replace the initial ${n}$ frames of the source video with the corresponding frames from the preceding clip $\#1$. Through this overlapping sampling and iterative inference method, our approach maintains consistency in the translation of videos with arbitrary lengths as shown in Fig.~\ref{fig:pipeline}(c).

\begin{table*}[]
\caption{Properties of different text-driven video editing methods. We compare the model type, additional control, tune time, and the application scopes. The reported tune time is from~\cite{InsV2V}. }
\vspace{-0.2cm}
\scalebox{0.76}{
\begin{tabular}{@{}ccccccccc@{}}
\toprule
Method           & ModelType         & Additional Ctrl  & \begin{tabular}[c]{@{}c@{}}Tune Time\\ /Video\end{tabular} & \begin{tabular}[c]{@{}c@{}}Need Latent\\  Inversion\end{tabular} & Prompt Type      & \begin{tabular}[c]{@{}c@{}}Support\\ Image-guided\end{tabular} & Multi-Task  & \begin{tabular}[c]{@{}c@{}}Support\\ Long Video\end{tabular} \\ \midrule
Tune-A-Video~\cite{tuneavideo}     & Per-Vid-Per-Model & No               & 15 mins                                                    & Yes                                                              & Original\&Target & No              & No                                                & No         \\
Vid2Vid-Zero~\cite{vid2vid-zero}     & Per-Vid-Per-Model & Prompt-to-Prompt & 12 mins                                                    & Yes                                                              & Original\&Target & No                                                             & No  & No        \\
Video-P2P~\cite{liu2023videop2p}        & Per-Vid-Per-Model & Prompt-to-Prompt & 10 mins                                                    & Yes                                                              & Original\&Target & No                                                             & No    & No      \\
ControlVideo~\cite{controlvideoone-shot}     & Per-Vid-Per-Model & ControlNet       & 15 mins                                                    & Yes                                                              & Original\&Target & No                                                             & No     & Yes     \\
SimDA~\cite{SimDA}     & Per-Vid-Per-Model & No       & 5 mins                                                    & Yes                                                              & Original\&Target & No                 & No                                             & No         \\
ReRender-A-Video~\cite{yang2023rerenderavideo}     & One-Model-All-Vid & ControlNet       & No Need                                                    & No                                                              & Original\&Target & No                                                             & No      & Yes    \\
Instruct-Vid2Vid~\cite{qin2023instructvid2vid} & One-Model-All-Vid & No               & No Need                                                    & No                                                               & Instruction      & No   & No                                                           & No         \\
InsV2V~\cite{InsV2V}           & One-Model-All-Vid & No               & No Need                                                    & No                                                               & Instruction      & No                                                             & No     & Yes     \\ \midrule
VIDiff(Ours)     & One-Model-All-Vid & No               & No Need                                                    & No                                                               & Instruction      & Yes                                 & Yes                             & Yes        \\ \bottomrule
\end{tabular}
}
\vspace{-0.3cm}
\label{tab:edit}
\end{table*}

\begin{table*}[]
\caption{Quantitative comparison with open-sourced evaluated baseline.  We report both the quantitative and the user study results. The ``Tuning" refers to the process of optimization. The ``Inference" time includes both Inversion and Denoising times.}
\vspace{-0.2cm}
\centering
\scalebox{0.75}{
\begin{tabular}{@{}c|ccc|cc|cc|cc@{}}
\toprule
\multirow{2}{*}{Method} & \multicolumn{3}{c|}{Video-Text Alignment} & \multicolumn{2}{c|}{Frame consistency} & \multicolumn{2}{c|}{Runtime {[}min{]}}  & \multirow{2}{*}{\begin{tabular}[c]{@{}c@{}}Support\\ Long Video\end{tabular}} & \multirow{2}{*}{\begin{tabular}[c]{@{}c@{}}Additional\\ Control\end{tabular}}      \\ \cmidrule(l){2-8} 
                        & CLIPScore($\uparrow$) & PickScore($\uparrow$)           & User Vote($\uparrow$)           & CLIPScore($\uparrow$)          & User Vote($\uparrow$)          & Tuning($\downarrow$)           & Inference($\downarrow$)       \\ \cmidrule(r){1-10}
Tune-A-Video~\cite{tuneavideo}            & 30.51  & 20.24                & 32.3\%                    &  91.21                & 30.5\%                   & 11.68              & 0.96   & No      & No         \\
Vid2Vid-Zero~\cite{vid2vid-zero}            &30.28  & 20.09                 &  30.1\%                   & 92.06                    & 29.6\%                   &  4.32                & 2.67        & No       & No     \\
ControlVideo~\cite{controlvideo}            &31.03   & \underline{20.57}                & 36.8\%                      & \bf{92.89}                 &  43.6\%                  & 3.75                 & 2.75         & Yes    & ControlNet       \\
Rerender-A-Video~\cite{yang2023rerenderavideo}        &\underline{31.09}    &  20.45                &  40.2\%                   & 90.87                 &  37.9\%                  & -                 &  4.12     & Yes   & ControlNet          \\ \midrule
VIDiff (Ours)           &  \bf 31.15    & \bf{20.73}            & -                   & \underline{92.20}                 & -                  & -                & \bf 0.54 & Yes    & No           \\ \bottomrule
\end{tabular}
}
\vspace{-0.4cm}
\label{tab:edit_result}
\end{table*}

\vspace{-0.1cm}
\section{Experiments}
\subsection{Settings}
\noindent\textbf{Dataset Details}
~~The model training is based on the triplet of $<$source, target, instruction$>$, which includes various video tasks mentioned above, such as video dehazing, deblurring, recoloring, inpainting, object segmentation, and video editing, \emph{etc}. Specifically, for \textbf{dehazing and deblurring} tasks, we utilized the HazeWorld~\cite{hazeworld} and BSD~\cite{BSD} datasets, respectively. For \textbf{video editing} tasks, we followed Instruct-Pix2Pix~\cite{brooks2023instructpix2pix}, using GPT-4~\cite{GPT4} and advanced video editing model~\cite{tuneavideo, vid2vid-zero, yang2023rerenderavideo} to create the triplet data. Due to limitations in the generality of current video editing models, we only generated 8,000 pairs of data, mainly focusing on style and color editing tasks. For the \textbf{video object segmentation} task, we constructed the training set using the DAVIS-RVOS~\cite{refer_davis} and Refer-YoutubeVOS~\cite{refer_youtube} datasets. During training, we formulated the target video as a semi-transparent mask. As for \textbf{video recoloring and inpainting} tasks, we believe any video dataset can train such networks. We employed the datasets mentioned above as well as part data from WebVid~\cite{webvid} for training these tasks.

\begin{figure}
    \centering
    \includegraphics[width=1.0\linewidth]{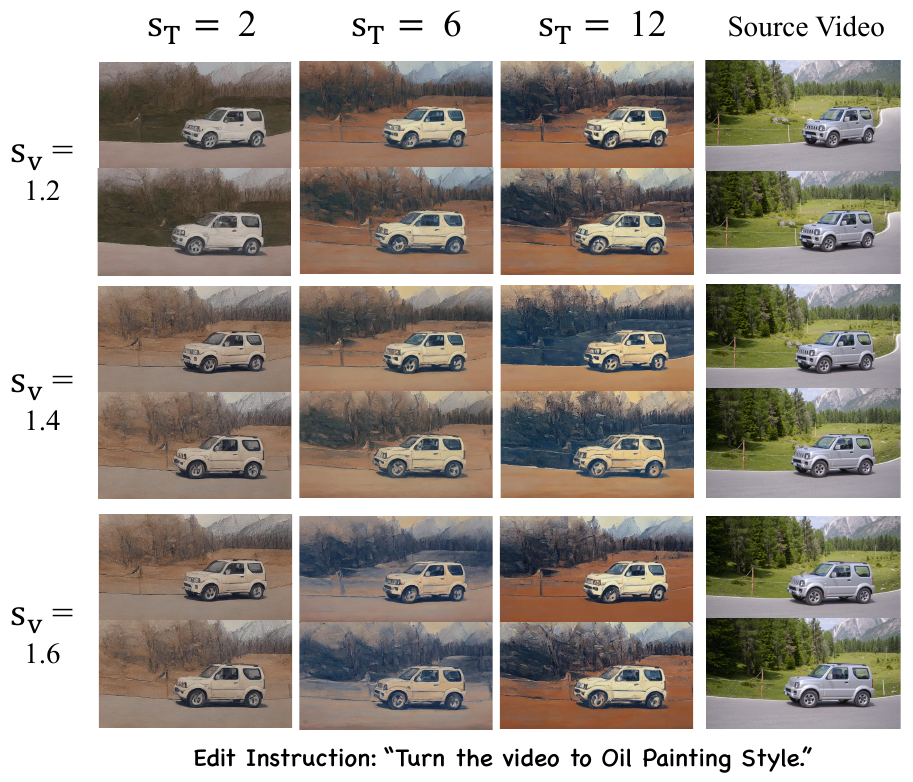}
    \vspace{-0.5cm}
    \caption{Classifier-free guidance weights over two conditional inputs. $s_V$ controls similarity with the input video, while $s_T$ controls
consistency with the edit instruction.}
    \label{fig:classifier-free}
    \vspace{-0.4cm}
\end{figure}

\vspace{0.2cm}
\noindent\textbf{Implementation Details}
~We use Stable Diffusion v1.5~\cite{stablediffusion} as initialization to leverage the text-to-image generation prior. Additionally, we utilize the motion module from AnimateDiff~\cite{AnimateDiff} to initialize the temporal layer for better temporal modeling. The learning rate during training is set to $1e-4$. The input video frames consist of 16 frames with a resolution of $256\times256$. We validate that even with low and fixed resolution during training, our approach can easily be extended to arbitrary resolutions and aspect ratios during inference. Once the training is completed, our method can be effortlessly applied to various video translation tasks. For each training step, we randomly select a task. In this way, we only need to train a unified model. Due to the mutual learning of multiple tasks, we confirm the effectiveness of the unified model. We employ classifier-free diffusion guidance~\cite{classifier-free} and introduce two guidance scales, $\text{s}_V$ and $\text{s}_T$. These scales can be adjusted to balance the degree to which the generated samples align with the input video and the extent to which they adhere to the editing instructions. We show the effects of these two parameters on generated samples in Fig.~\ref{fig:classifier-free}.

\subsection{Experimental Results}
In this section, we will compare our model with baseline methods across multiple different tasks.

\vspace{0.2cm}
\noindent\textbf{Video Editing}
~~We first list several common video editing baseline methods in Table~\ref{tab:edit}, comparing their various attributes. It can be seen that most of these methods are based on one-shot tuning, meaning they require training a dedicated model for the specific video to be edited. This not only requires additional tuning time but also detailed textual descriptions of both the source and target videos, greatly limiting their generality. Our approach is similar to~\cite{qin2023instructvid2vid, InsV2V} in that it does not require additional tuning. Additionally, our method supports multi-modal instructions, various tasks, and the editing of long videos.

We also follow the previous benchmark methods, replicating several open-source video editing techniques for comparison. We report CLIP Score~\cite{clip}, PickScore~\cite{pickscore}, and Frame Consistency between different frames following~\cite{editbenchmark}. Additionally, we conducted a user study in which participants were presented with two sets of reports: one from our method and one from other methods. They were asked to choose the one with better matching of text and video as well as smoother video continuity. The experimental results are shown in Table~\ref{tab:edit_result}. We also report the tuning and inference time on a single NVIDIA A100 GPU.

\vspace{0.2cm}
\noindent\textbf{Video Re-colorization}
~~Regarding the video recoloring task, we conduct the experiment on a widely used benchmark. We followed previous studies and validated our approach on the validation set of the DAVIS~\cite{davis} dataset. Evaluating video recoloring tasks generally involves assessing perceptual realism, color vividness, and temporal consistency. To evaluate the perceptual realism of colorized videos, we used the FID~\cite{fid} (Fréchet Inception Distance) metric, which measures the similarity between the predicted colors and the ground truth distribution. To assess color vividness, we employed the Colorfulness~\cite{colorfulness} metric. Additionally, to evaluate temporal consistency, we utilized the Color Distribution Consistency~\cite{CDC}(CDC) index. Furthermore, we also reported metrics such as PSNR~\cite{PSNR}, SSIM~\cite{SSIM}, and LPIPS~\cite{LPIPS} in our study.
We compare our method with several automatic video colorization techniques as well as exemplar-based video colorization baselines. The quantitative results are presented in Table ~\ref{tab:color}, demonstrating significant improvements in our approach based on perceptual evaluation metrics. Moreover, our method maintains a comparable performance in structural metrics.

\begin{table}[]
\caption{Quantitative results on the evaluation datasets from different methods. The best items and second best items are highlighted in bold and underlined respectively. }
\vspace{-0.2cm}
\tabcolsep=0.1cm
\scalebox{0.72}{
\begin{tabular}{@{}lcccccc@{}}
\toprule
Method        & FID($\downarrow$)   & Color($\uparrow$)        & PSNR($\uparrow$)   & SSIM($\uparrow$)   & LPIPS($\downarrow$) & CDC($\downarrow$)      \\ \midrule
AutoColor~\cite{autocolor}     & 83.05 & 14.14        & \underline{24.41} & 0.915 & 0.264 & 0.003734 \\
Deoldify~\cite{deoldify}      & 76.21 & 25.47        & 23.99 & 0.885 & 0.306 & 0.004901 \\
DeepExemplar~\cite{deepexemplar}  & 77.26 & 28.82        & 21.78 & 0.846 & 0.325 & 0.004006 \\
DeepRemaster~\cite{deepremaster}  & 97.54 & 25.66        & 21.95 & 0.848 & 0.354 & 0.005098 \\
TCVC~\cite{tcvc}          & 74.94 & 21.72        & \bf 25.17 & \underline{0.921} & 0.239 & \underline{0.003649} \\
VCGAN~\cite{vcgan}         & 70.29 & 15.89        & 23.90 & 0.910 & 0.247 & 0.005303 \\
ColorDiffuser~\cite{colorDiffusers} &\underline{69.51} & \underline{29.13}        & 23.73 & \bf 0.939 & \underline{0.213} & \bf 0.003607 \\ \midrule
VIDiff (Ours) & \bf 63.96 & \bf 32.84        & 23.59 & 0.895 & \bf 0.196      & 0.003994         \\ \bottomrule
\end{tabular}
}
\label{tab:color}
\vspace{-0.5cm}
\end{table}

\begin{table*}[]
\caption{ Quantitative results on video enhancement. We also report an upper baseline,  obtained by reconstructing the ground truth images using VAE, representing the performance upper bound achievable with the used VAE model.  }
\vspace{-0.3cm}
\scalebox{0.75}{
\begin{tabular}{@{}c|cccc|cccc|cccc@{}}
\toprule
\multirow{2}{*}{Method} & \multicolumn{4}{c|}{Deblurring}                         & \multicolumn{4}{c|}{Dehazing} & \multicolumn{4}{c}{In-painting} \\ \cmidrule(l){2-13} 
                        & PSNR($\uparrow$) & FID($\downarrow$) & LPIPS($\downarrow$) & NIQE($\downarrow$) & PSNR($\uparrow$)     & FID($\downarrow$)    & LPIPS($\downarrow$) & NIQE($\downarrow$)     & PSNR($\uparrow$)      & FID($\downarrow$)     & LPIPS($\downarrow$)  & NIQE($\downarrow$)   \\ \cmidrule(r){1-13}
Instruct-Pix2Pix~\cite{brooks2023instructpix2pix}        &  19.51                & 61.71       & 0.4260       & \underline{12.65}         & 14.24        & 89.29         &0.6375   &  14.05             &15.44           & 82.07        & 0.3313     &\underline{24.94}      \\

MagicBrush~\cite{magicbrush}              & 23.45        &  38.05               &0.2977        &14.16          &15.13         & \underline{27.60}         & 0.5314    & 14.20          & 16.15          & 56.14        & 0.2659       & 25.74  \\
Instruct Diffusion~\cite{instructdiffusion}      &   \underline{25.62}                     & \bf 14.05      & \underline{0.1775}          & 14.66        &\underline{16.72}          & 32.73    &  \underline{0.5188}   & \underline{13.97}         & \underline{20.85}          & \underline{38.01}        & \underline{0.2015}         &31.37 \\ \midrule
VIDiff (Ours)                    & \bf 27.68                    & \underline{14.17}  &\bf 0.1633         &\bf 10.12   &\bf 20.68    &\bf 19.55   &\bf 0.1319      &\bf 10.42     &\bf 23.17          &\bf 19.21        &\bf 0.1314      &\bf 24.33   \\ 
VAE Recon (Upper)                    &  29.15                   & 4.782  & 0.0587         &13.25    & 27.34    & 6.313  & 0.0908  & 10.79  &  25.54        & 8.981                 &0.0847      & 26.54    \\ 
\bottomrule
\end{tabular}
}
\vspace{-0.2cm}

\label{tab:enhancement}
\end{table*}

\begin{figure*}
    \centering
    \includegraphics[width=1.0\textwidth]{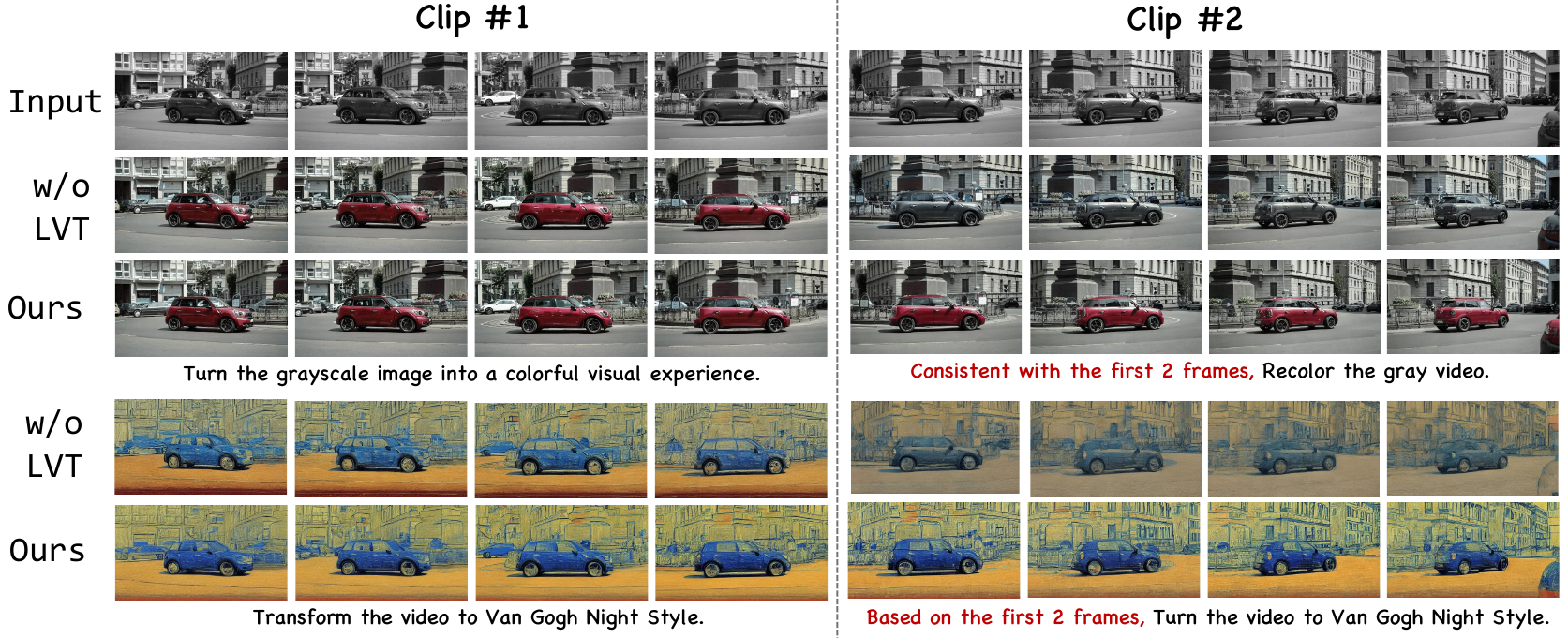}
    \vspace{-0.5cm}
    \caption{Ablation of our auto-regressive long video translation. VIDiff could maintain temporal consistency across different video clips.}
    \label{fig:ablation_lvt}
    \vspace{-0.4cm}
\end{figure*}

\vspace{0.2cm}
\noindent\textbf{Video Enhancement}
~~We evaluate the performance of our model on video enhancement tasks across several common benchmarks. We follow the evaluation method as~\cite{BSD,hazeworld,propainter}, which includes evaluation datasets commonly used in this field such as the test set of BSD~\cite{BSD}, Youtube~\cite{refer_youtube}, and DAVIS~\cite{davis}, among others.  The quantitative results are presented in Table~\ref{tab:enhancement}. We not only report the distortion metric PSNR~\cite{PSNR} to measure the difference between the edited frame and the ground truth, but also follow the method described in~\cite{multiscalediffusion} to calculate some aesthetic perceptual image quality metrics such as FID~\cite{fid}, LPIPS~\cite{LPIPS}, and NIQE~\cite{niqe}. We compare our method with several mainstream open-source instructive editing techniques. It can be observed that our method outperforms others significantly across all metrics. Lastly, the performance of our model in image enhancement is constrained by the VAE~\cite{VQVAE} model, which introduces information loss. Therefore, we also report the results of VAE reconstructing the original image and compare it with the ground truth. This serves as an upper baseline, allowing us to measure the upper limit that methods based on the LDM~\cite{stablediffusion} can achieve.

\noindent\textbf{Visualizations}
Here we provide more visualization results for the VIDiff method of video translation. We present results of  the video re-colorization task in Fig.~\ref{fig:recolor}, video dehazing and video in-painint in Fig.~\ref{fig:haze_painting}, video deblurring and language-guided video object segmentation in Fig.~\ref{fig:rvos}. We also show the multi-modal instruction guided video editing in Fig.~\ref{fig:editing}. For fully rendered videos, we primarily refer the reader to our project page (\url{https://chenhsing.github.io/VIDiff}).

\begin{figure}
    \centering
    \includegraphics[width=1.0\linewidth]{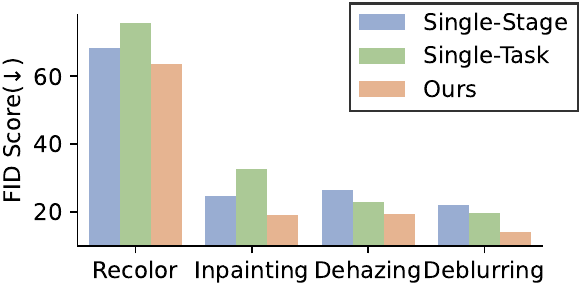}
    \vspace{-0.6cm}
    \caption{Ablation study on multi-task training and multi-stage transferring method. We evaluate our models on four tasks. It demonstrates that joint multi-task training and multi-stage transferring significantly enhance the performance of each task.}
    \vspace{-0.3cm}
    \label{fig:ablation}
\end{figure}

\subsection{Ablation Study}
\label{Sec:ablationstudy}
\noindent\textbf{The Effectiveness of Multi-task Training}
~~Presently, multi-task learning has become increasingly popular. It not only allows a single model to handle multiple related tasks but also enables the model to achieve better generalization performance. We conduct experiments to compare our multi-task learning model with individually trained single-task models. The performance differences are reported in Fig.~\ref{fig:ablation}. This comparison was made across four task-specific test datasets, demonstrating that our jointly trained model outperforms the specialized models. Clearly, the model trained jointly performs better. Additionally, we observed that this advantage extends to the field of video editing. We will present more qualitative comparison results in the supplementary materials.

\vspace{0.2cm}
\noindent\textbf{The Benefit of Multi-Stage Transferring Learning}
~~As we know, most video editing methods rely on transferring pre-trained T2I model Stable Diffusion~\cite{stablediffusion}. However, approaches like Instruct-Vid2Vid~\cite{qin2023instructvid2vid} and InsV2V~\cite{InsV2V} directly transfer T2I to V2V. The original model lacks motion information, resulting in poor temporal modeling. Fine-tuning makes the model focus more on temporal modeling, neglecting the spatial transfer inherent in the model. We employed a multi-stage training approach that effectively mitigates this issue. Our ablation experiments, as illustrated in Fig.~\ref{fig:ablation}, demonstrated that the multi-stage transfer method yields significantly better image quality compared to direct fine-tuning.

\vspace{0.2cm}
\noindent\textbf{The Effectiveness of Long Video Translation}
~~For tasks such as video re-colorization and video editing, ensuring consistency in long videos is a crucial challenge. Methods based on diffusion are primarily trained on short video clips~\cite{tuneavideo, SimDA, controlvideoone-shot}, and therefore, they can only edit relatively short video clips. The auto-regressive long video translation paradigm we propose effectively addresses this issue. In our comparative experiment shown in Fig.~\ref{fig:ablation_lvt}, it can be observed that a given long video, is typically divided into different clips. Without the Long Video Translation (LVT) constraints, these different clips are entirely inconsistent with each other. With our approach, the method can maintain excellent consistency between different clips in long videos.

\section{Conclusion}
In conclusion, this paper introduced Video Instruction Diffusion (VIDiff), a novel unified framework for aligning video tasks with human instructions. VIDiff treated various video understanding tasks as conditional video translation problems, allowing us to translate videos into desired outcomes based on instructions. We demonstrated the effectiveness of our approach across multiple tasks, with joint training enhancing the generalization capabilities of the model. This research marked a significant step in constructing a universal modeling interface for video tasks, paving the way for future advancements in the pursuit of artificial general intelligence in video understanding. In future work, we plan to further explore the performance and capabilities of VIDiff, considering potential integration with large language models to enable more versatile unified video tasks such as video question answers and video contextual understanding.

{
    \small
    \bibliographystyle{ieeenat_fullname}
    \bibliography{main}
}

\begin{figure*}[h]
\centering
\includegraphics[width=0.92 \linewidth]{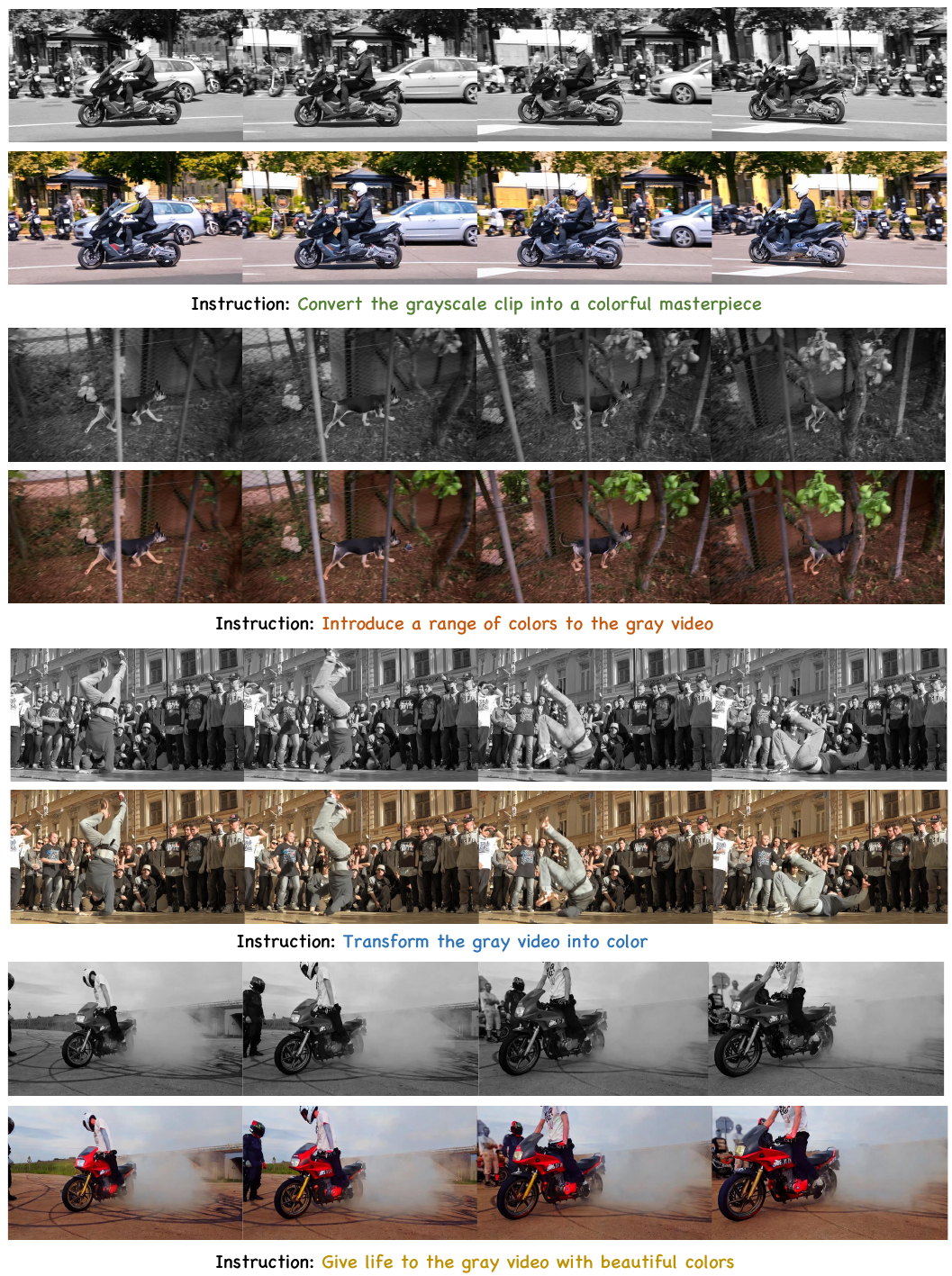} 
\caption{Results of extending our VIDiff to video re-colorization task.
}
\label{fig:recolor}
\end{figure*}

\begin{figure*}[h]
\centering
\includegraphics[width=0.91\linewidth]{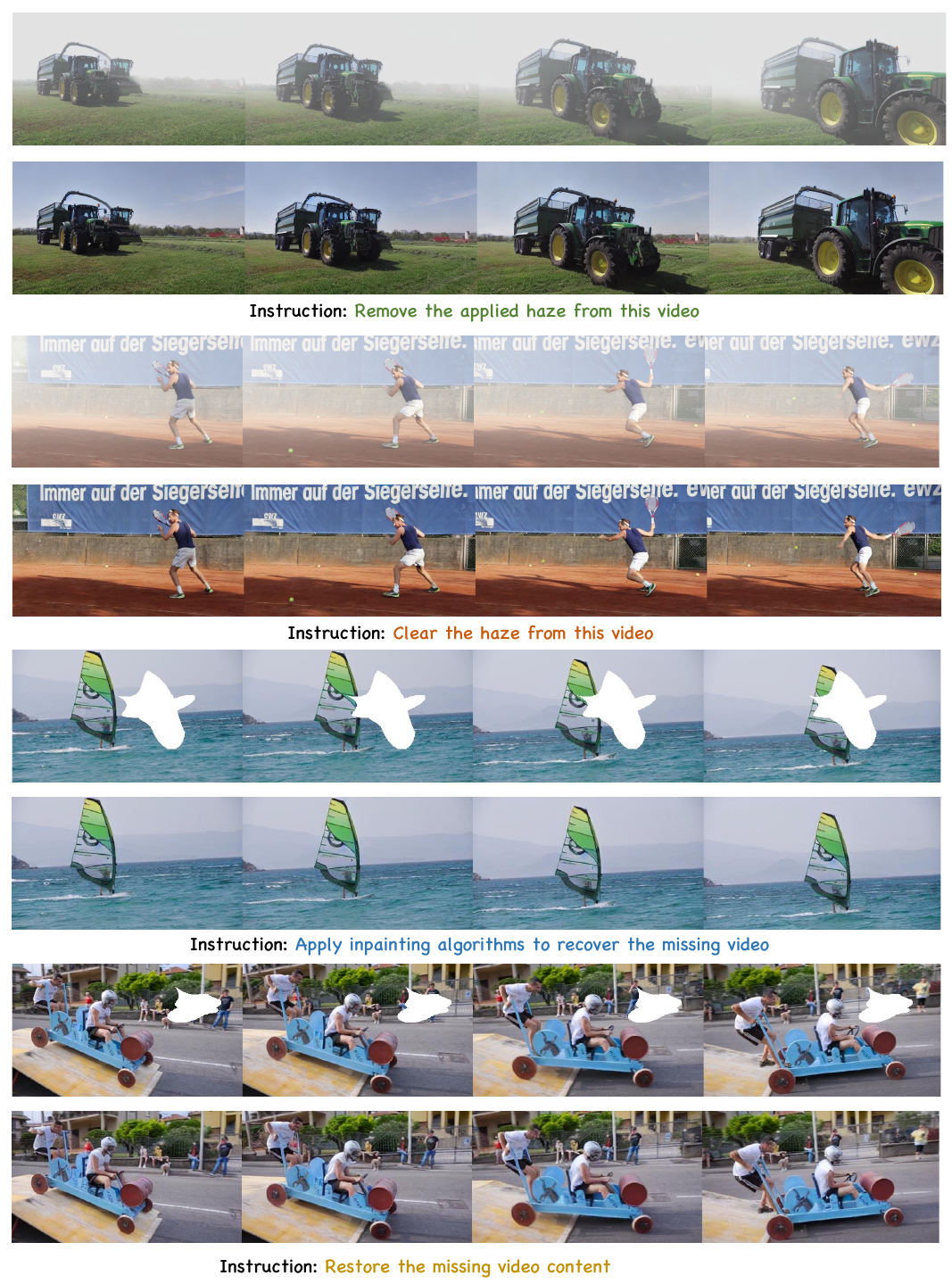} 
\caption{Results of extending our VIDiff to video dehazing and in-painting task.
}
\label{fig:haze_painting}
\end{figure*}

\begin{figure*}[h]
\centering
\includegraphics[width=0.91\linewidth]{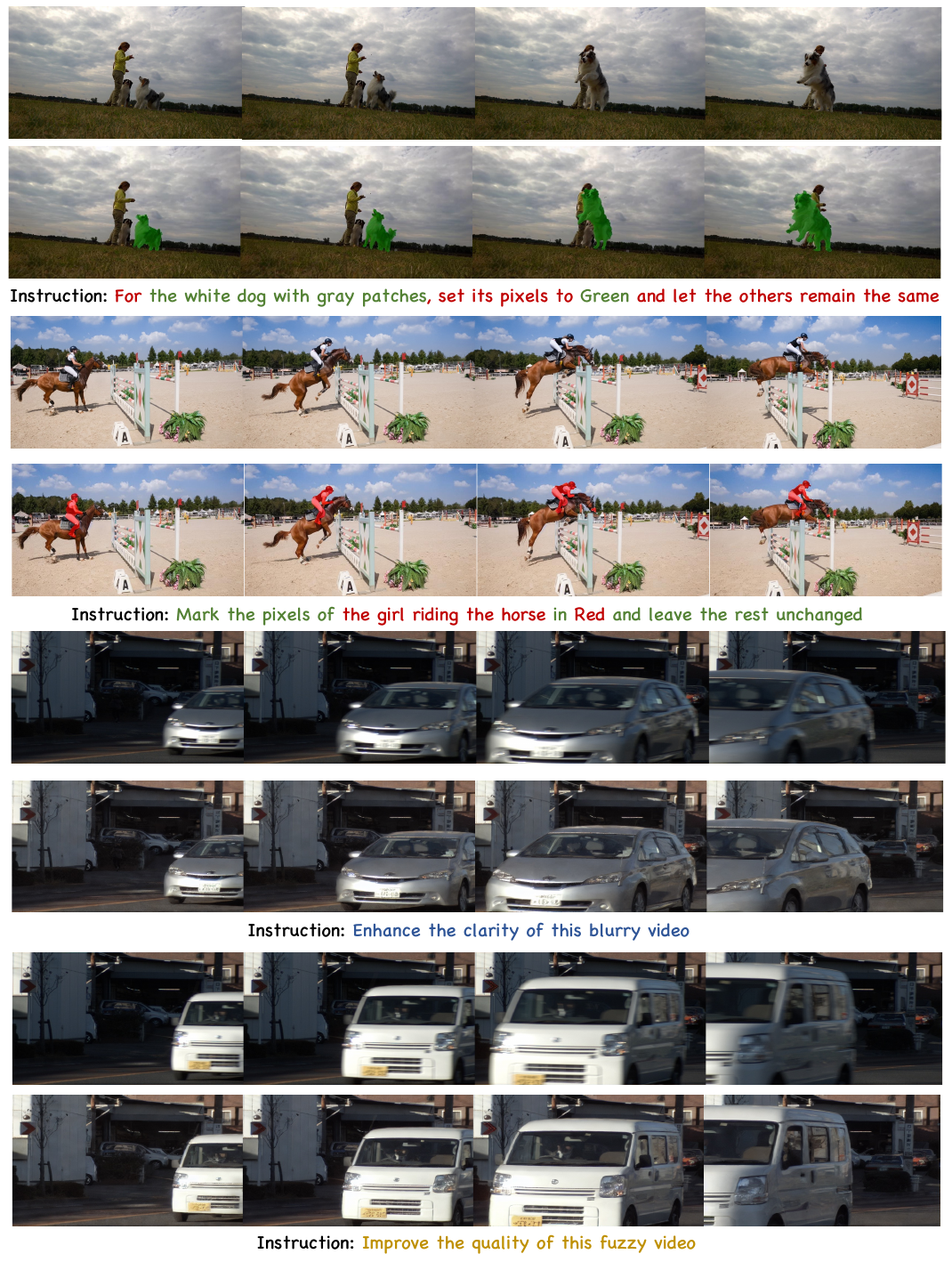} 
\caption{Results of extending our VIDiff to language guided video object segmentation  and deblurring task.
}
\label{fig:rvos}
\end{figure*}

\begin{figure*}[h]
\centering
\includegraphics[width=0.93\linewidth]{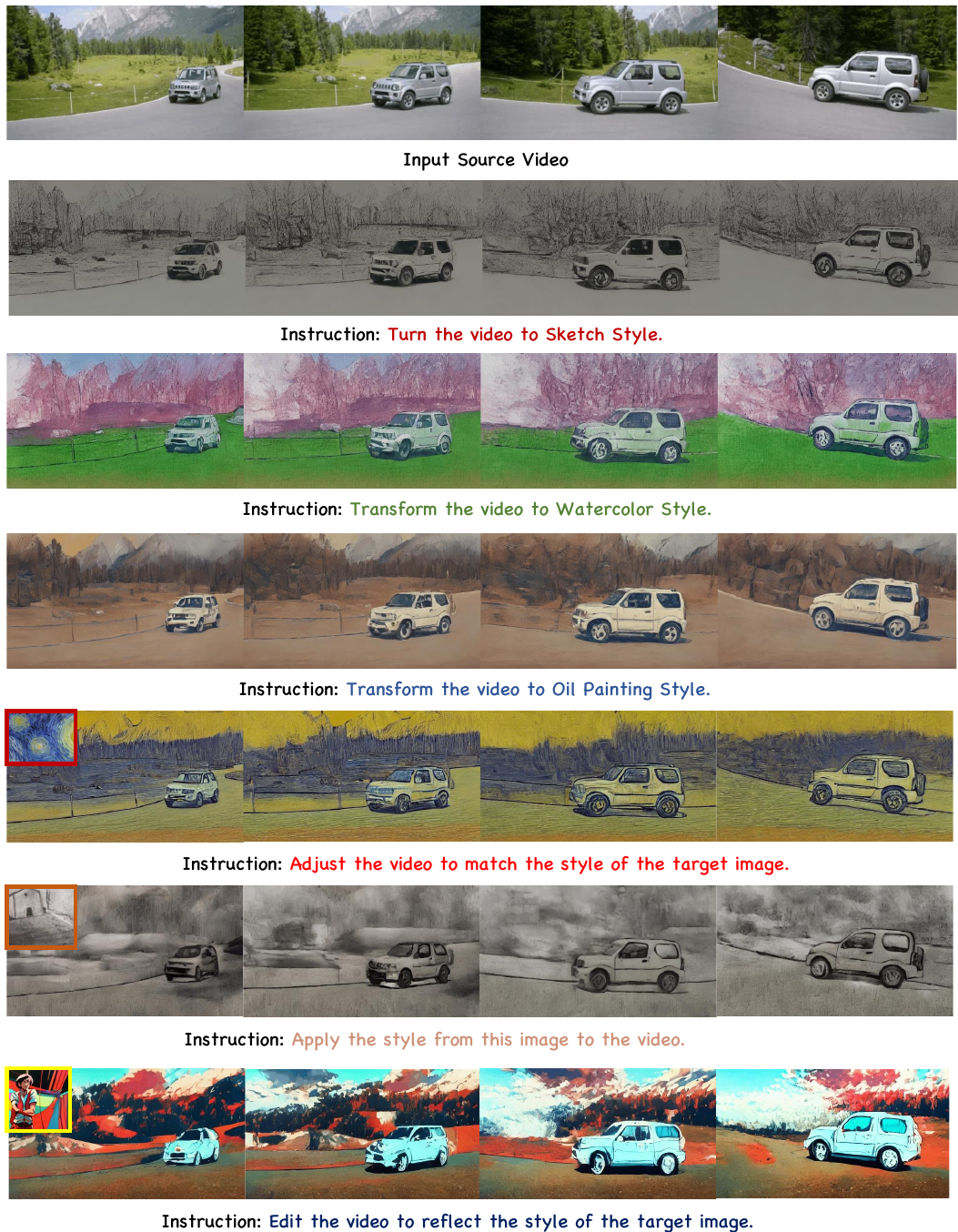} 
\caption{Results of extending our VIDiff to video editing task with both single-modal and multi-modal instructions.
}
\label{fig:editing}
\end{figure*}

\end{document}